\pdfoutput=1

\documentclass[11pt]{article}

\usepackage{hyperref}

\usepackage[final]{acl}

\usepackage{times}
\usepackage{latexsym}

\usepackage[normalem]{ulem} 
\usepackage{soul} 

\usepackage[T1]{fontenc}

\usepackage[utf8]{inputenc}

\usepackage{microtype}

\usepackage{inconsolata}

\usepackage{graphicx}

\usepackage{booktabs}
\RequirePackage{makecell}
\usepackage{xcolor, colortbl}
\usepackage{comment}

\usepackage{booktabs}
\usepackage{multirow}

\usepackage{graphicx}
\usepackage{subcaption}

\usepackage[arabic,farsi,english]{babel}

\newcommand{\blue}[1]{{\color{blue}#1}}

\newcommand{\cyan}[1]{{\color{cyan}#1}}
\newcommand{\green}[1]{{\color{green}#1}}
\newcommand{\orange}[1]{{\color{orange}#1}}
\newcommand{\purple}[1]{{\color{purple}#1}}

\title{Comparative Study of Multilingual Idioms and Similes \\ in Large Language Models}

\author{
 \textbf{Paria Khoshtab\thanks{Equal contribution, ordered alphabetically.}\textsuperscript{1}  }
 \textbf{   Danial Namazifard\footnotemark[1]\textsuperscript{1}    }
\\
 \textbf{Mostafa Masoudi\textsuperscript{1, 2}  }
 \textbf{       Ali Akhgary\textsuperscript{1}  }
 \textbf{       Samin Mahdizadeh Sani\textsuperscript{1}    }
\\
 \textbf{Yadollah Yaghoobzadeh\textsuperscript{2, 1}}
\\
 \textsuperscript{1}School of Electrical and Computer Engineering\\
 College of Engineering University of Tehran, Tehran, Iran \\
 \textsuperscript{2}Tehran Institute for Advanced Studies, Khatam University, Tehran, Iran
\\
{\small\texttt{\{paria.khoshtab, namazifard, mostafa.masoudi, ali.akhgary, samin.mehdizadeh, y.yaghoobzadeh\}@ut.ac.ir}}
}

\begin{document}
\maketitle
\begin{abstract}
This study addresses the gap in the literature concerning the comparative performance of LLMs in interpreting different types of figurative language across multiple languages. By evaluating LLMs using two multilingual datasets on simile and idiom interpretation, we explore the effectiveness of various prompt engineering strategies, including chain-of-thought, few-shot, and English translation prompts. We extend the language of these datasets to Persian as well by building two new evaluation sets. Our comprehensive assessment involves both closed-source (GPT-3.5, GPT-4o mini, Gemini 1.5), and open-source models (Llama 3.1, Qwen2), highlighting significant differences in performance across languages and figurative types. 
Our findings reveal that while prompt engineering methods are generally effective, their success varies by figurative type, language, and model. 
We also observe that open-source models struggle particularly with low-resource languages in similes. Additionally, idiom interpretation is nearing saturation for many languages, necessitating more challenging evaluations.\footnote{Data and code: \href{https://github.com/namazifard/Multilingual-Idioms-Similes}{https://github.com/namazifard/Multilingual-Idioms-Similes}}
\end{abstract}

\section{Introduction}
Large language models (LLMs) have revolutionized NLP by demonstrating remarkable capabilities in understanding and generating human language.
One of the most challenging aspects of human language for LLMs to comprehend is figurative language, which includes similes, idioms, and metaphors.
Figurative language significantly enriches human communication by facilitating the implicit expression of complex ideas and emotions \cite{roberts1994people, fussell2014figurative}. Unlike literal expressions, figurative language often involves rich cultural references and judgments that vary considerably across different cultures \cite{shutova2011computational, fussell2014figurative}. Consequently, understanding and generating figurative language is crucial for LLMs to interact naturally and effectively with users. Therefore, studying how these models handle figurative language is essential for advancing their capabilities.

Recent studies have highlighted that LLMs,
not only struggle to generate but also frequently misinterpret figurative expressions \cite{huang2024survey}, underscoring the need for more sophisticated techniques to bridge these gaps. The challenge becomes even more pronounced in multilingual contexts, where figurative language is intricately tied to cultural nuances \cite{liu-etal-2024-multilingual}.

There remains a gap in the literature regarding the comparative performance of LLMs in interpreting different types of figurative language, in English and multilingual contexts. 
This study focuses on two types of figurative language: similes and idioms. 
A simile compares two entities, typically using ``like'' or ``as'' (e.g., ``as busy as a bee''), to create vivid descriptions. An idiom, by contrast, is a fixed phrase whose meaning cannot be inferred from the meanings of its components (e.g., ``kick the bucket'').
These two are distinct in their structure and usage, suggesting that LLMs might perform differently on them, and require different strategies to process them effectively.  

We evaluate the performance of multiple LLMs across various languages using two existing datasets: MABL \cite{kabra-etal-2023-multi}, MAPS \cite{liu-etal-2024-multilingual}, and our newly developed Persian datasets for simile and idiom interpretation. MABL includes examples of figurative language interpretation as an inference task, and mainly simile expressions. It covers eight languages in high and low resource ranges. MAPS is a multilingual dataset of proverb interpretation including six languages.
To further advance this research, we contribute by extending the scope to Persian, the native language of the authors, by developing two additional evaluation sets. These new sets help us analyze the datasets and model performance more deeply.

Evaluating LLMs requires interacting with them, making prompt engineering a critical component for optimizing performance.
We examine various prompting strategies, including chain-of-thought, few-shot, and dialogue simulation. We extend our evaluations to input being translated to English, as it serves as a strong baseline for many multilingual evaluations \cite{lin-etal-2022-shot,shi2023language,liu-etal-2024-multilingual}.
To achieve a comprehensive evaluation, we conduct an exhaustive assessment using both closed-source—GPT-3.5, GPT-4o mini, and Gemini 1.5 \cite{geminiteam2024gemini15unlockingmultimodal}—and open-source Llama 3.1 \citep[8B, 70B]{dubey2024Llama3herdmodels} and Qwen2 \citep[7B, 72B]{yang2024qwen2technicalreport}.

Our findings reveal several novel insights:
(i) Prompt engineering methods show varying degrees of success depending on the figurative type, language, and model. 
(ii) Open-source models perform similarly to closed-source models in idioms, but they generally lag behind in interpreting similes. 
(iii) The interpretation of idioms in the style of the MAPS dataset is nearing saturation for many languages when strong LLMs are used, due to the presence of idioms and their meanings in training data. (iv) The presentation of pre-trained data, as well as the script used in different languages, significantly impact model performance. (v) Chain-of-thought prompting proves particularly effective for simile interpretation in smaller models.

\section{Related work}
\subsection{Figurative language processing}
Figurative expressions encapsulate complex human experiences and cultural knowledge, making them essential in tasks ranging from sentiment analysis \cite{hercig-lenc-2017-impact} to machine translation \cite{wang2024mmte}. Previous research efforts have focused on various types of figurative language, with several approaches dedicated to improving simile detection and component extraction \cite{qadir-etal-2015-learning, mpouli-2017-annotating, liu-etal-2018-neural, zeng2019neuralsimilerecognitioncyclic}, as well as on generating similes \cite{chakrabarty-etal-2020-generating, zhang2021writing, lai-nissim-2022-multi}. 

In addition to similes, proverbs and idioms are another type of figurative expression that has been explored in various studies,  focusing on identifying whether a phrase is used idiomatically or proverbially, either within a specific context (token-level) or in general (type-level) \cite{li-sporleder-2009-classifier, fazly-etal-2009-unsupervised, verma-vuppuluri-2015-new, salton-etal-2016-idiom, peng-feldman-2016-experiments}. Beyond detection and generation, researchers have examined methods for interpreting and representing these figurative expressions, including literal paraphrasing, treating them as single tokens, or composing them from characters rather than words \cite{liu-hwa-2016-phrasal, zhou-etal-2021-pie}. 
In a concurrent study on idiom translation, \citet{rezaeimanesh2024comparative} introduce the PersianIdioms dataset—comprising parallel Persian-English and English-Persian samples, designed for evaluating idiom translation—and compare the performance of LLMs, NMT systems, and hybrid methods.

Additionally, significant research has been conducted on other types of figurative language, such as metaphors, sarcasm, and irony \cite{yu-wan-2019-avoid, ghosh-etal-2017-role, chakrabarty-etal-2021-mermaid}.
Additionally, significant research has been conducted on other types of figurative language, such as metaphors, sarcasm, and irony \cite{yu-wan-2019-avoid, ghosh-etal-2017-role, chakrabarty-etal-2021-mermaid}.

While many studies have focused on multilingual figurative language detection \cite{lai-etal-2023-multilingual, tedeschi-etal-2022-id10m, tayyar-madabushi-etal-2022-semeval,aghazadeh-etal-2022-metaphors} and English multi-figurative (multiple types of figurative) language processing \cite{jhamtani-etal-2021-investigating, chakrabarty-etal-2022-rocket}, our research centers on multilingual multi-figurative language interpretation, which is a highly understudied area.

Our work expands upon the foundation laid by \citet{liu-etal-2024-multilingual} by introducing both similes and idioms across a wider range of languages, including the creation of new Persian datasets for each figurative type. While \citet{liu-etal-2024-multilingual} focus exclusively on proverbs and sayings, we extend the analysis to similes, offering a broader evaluation of LLMs' figurative language comprehension, and a comparative study of them.

\subsection{Multilingual prompt engineering} 
While LLMs have achieved impressive success in various NLP tasks \cite{NEURIPS2020_1457c0d6}, they encounter significant challenges in tasks that require understanding culturally specific figurative language \cite{li2024cultureparkboostingcrossculturalunderstanding}. The high cost of collecting multilingual cultural data further complicates these tasks. As a result, current methods to enhance the cultural awareness of LLMs rely primarily on two approaches: prompt engineering and culture-specific pre-training \cite{li2024culturellmincorporatingculturaldifferences}.

One prominent strategy is \textbf{Chain-of-Thought (CoT)} prompting, which has been demonstrated to improve LLM performance on various reasoning tasks by breaking down complex problems into more manageable steps \cite{wei2023chainofthoughtpromptingelicitsreasoning}. \citet{shi2023language} highlight the effectiveness of multilingual CoT prompting on reasoning benchmarks, though their evaluation does not include figurative language interpretation, such as similes and idioms, which is central to our work. We extend this research by applying CoT prompting specifically to simile interpretation, where cultural reasoning is often required.
Another promising technique is \textbf{ Reasoning in Conversation (RiC)}, introduced by \citet{wang-etal-2024-reasoning}, which simulates dialogue to improve performance on subjective, culturally-related tasks. In our study, we apply RiC specifically to simile tasks, leveraging the conversational context to enhance model reasoning for these culturally nuanced expressions.

In addition to prompt engineering, \textbf{translate-test} methods are commonly employed to address multilingual challenges. In this approach, evaluation data is translated into English before processing, using tools like Google Translate or LLMs \cite{ahuja-etal-2023-mega, liu2024translationneedstudysolving, shi2023language}. This method has proven effective in reducing performance gaps across languages \cite{conneau-etal-2018-xnli, ponti-etal-2020-xcopa, artetxe-etal-2023-revisiting}.
Building on this, \citet{huang-etal-2023-languages} propose Cross-Lingual-Thought (XLT) prompting, that translates the question into English and solves the problem in English before generating a response in the original language. We study the impact of English translation on simile and idiom interpretation across several languages, evaluating the consistency of responses between original and translated inputs under zero-shot and CoT settings.

\section{Methodology}
We focus on the task of figurative language interpretation, specifically across two types of figurative expressions: similes and idioms. Additionally, we extend the evaluation to Persian by creating new test sets for this language. Below, we describe the datasets used in our study, followed by an overview of the LLMs and prompting techniques applied.

\subsection{Datasets}
We employ two existing datasets: MABL (Metaphors Across Borders and Languages) \cite{kabra-etal-2023-multi} for simile experiments, and MAPS (Multicultural Proverbs and Sayings) \cite{liu-etal-2024-multilingual} to assess idioms. Both datasets facilitate the analysis of figurative language understanding across multiple languages, providing a diverse multilingual resource. A description of the datasets can be found in Appendix~\ref{appendix:datasets_intro}; however, key points are explained in the following.

\paragraph{MABL} contains figurative expressions in eight languages: English, Indonesian, Hindi, Swahili, Yoruba, Kannada, Sundanese, and Javanese. This dataset captures cultural and linguistic diversity in figurative language, offering a valuable resource for testing multilingual LLMs' abilities. We randomly select 200 simile samples (to be close to the number of examples in the idiom dataset, ensuring balance and comparability between the two datasets) from each language for evaluation, as shown in Table~\ref{tab:mabl_dataset}.

\begin{table}[htbp]
    \centering
    \resizebox{7.5cm}{!}{
        \begin{tabular}{p{3cm}|l|l|c}
            \toprule
            \textbf{start phrase} & \textbf{ending 0} & \textbf{ending 1} & \textbf{label} \\
            \midrule
            The test is as easy as rocket science & The test is easy & The test is hard & 1 \\
            \bottomrule
        \end{tabular}
    }
    \caption{An English example from MABL.}

    \label{tab:mabl_dataset}
\end{table}

\paragraph{MAPS} consists of proverbs and sayings, designed to evaluate their interpretation within conversational contexts. The dataset provides binary labels, indicating whether the proverb is used figuratively. It spans six languages: English, Indonesian, Mandarin Chinese, Bengali, German, and Russian, with sample counts of 214, 267, 143, 272, 183, and 226 respectively. We select idiomatic sentences for evaluation, as detailed in Table~\ref{tab:maps_dataset}.

\begin{table}[htbp]
    \centering
    \resizebox{7.5cm}{!}{
        \begin{tabular}{p{2.5cm}|p{3cm}|p{2.5cm}|p{2.5cm}|c}
            \toprule
            \textbf{proverb} & \textbf{conversation} & \textbf{answer A} & \textbf{answer B} & \textbf{label} \\
            \midrule
            fair exchange is no robbery & Person 1: ``Can I borrow your pen?'' Person 2: ``Sure, can I borrow your notebook?'' Person 1: ``Fair exchange is no robbery'' & Person 1 will lend the notebook to Person 2. & Person 1 will not lend the notebook to Person 2. & A \\
            \bottomrule
        \end{tabular}
    }
    \caption{An English example from MAPS.}
    \label{tab:maps_dataset}
\end{table}

\paragraph{Persian datasets} In addition to using the existing datasets, we created two new datasets specifically for the Persian language, following the formats of MABL and MAPS for similes and idioms that provide resources for evaluating figurative language understanding in Persian, which is underrepresented in current multilingual model researches.

\textbf{(i) Persian simile} We follow the methodology used in the MABL dataset but utilize GPT-4o to assist in generating examples. The model generates simile pairs by producing a start phrase with two possible endings—one reflecting the correct simile interpretation and the other conveying an incorrect meaning. The Persian Simile dataset consists of 200 samples. The prompts closely follow the instructions from the MABL dataset, ensuring consistency with the format and objectives of the original dataset (Appendix~\ref{appendix:dataset-construction}). After generation, three native Persian speakers manually evaluate and correct the examples to ensure both accuracy and cultural relevance.

\textbf{(ii) Persian idiom} We follow a methodology inspired by the creation process of the original MAPS dataset. First, we collect Persian idioms from two online resources: Daneshchi \footnote{\href{https://www.daneshchi.ir/}{https://www.daneshchi.ir}}, an educational portal, and Abadis \footnote{\href{https://abadis.ir/}{https://abadis.ir}}, an online dictionary. To create conversational contexts based on each idiom's explanation, we develop prompts using examples written by native Persian speakers to guide the model in understanding how idioms are used in everyday conversation. Using GPT-4o, we generate a conversational context for each idiom, ensuring that the idiom is correctly situated in a natural dialogue. In the second round, we provide the model with the idiom, explanation, and the generated conversational context. The model is then tasked with generating two response choices—one correct and one incorrect—based on the meaning of the idiom in the conversation. The Persian Idiom dataset contains 316 samples. Finally, native speakers review the generated content to ensure its accuracy, cultural appropriateness, and grammatical correctness.

For more details on the dataset construction and verification process, see Appendix~\ref{appendix:dataset-construction}.

\subsection{Language Categorization}
\begin{figure}[!ht]
    \centering
    \includegraphics[width=1\linewidth]{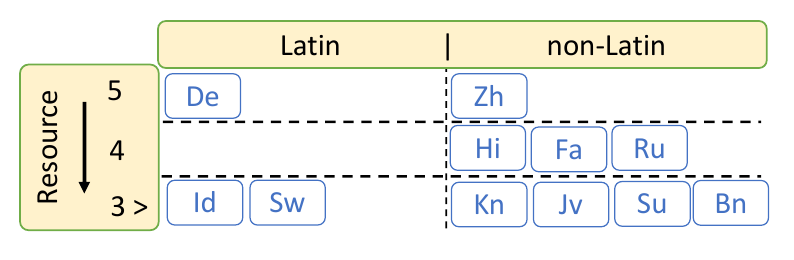}
    \caption{Our language categorization provides clearer insights for analyzing the results.}
    \label{fig:lang_categorization}
\end{figure}

Based on our experiment results we propose a categorization of languages from two perspectives, as shown in Figure~\ref{fig:lang_categorization}. The script of a language plays a significant role, as Latin-based languages like English dominate the pretraining data of large language models. Additionally, \citet{joshi-etal-2020-state} classify languages into six categories based on the availability of labeled and unlabeled data on the web. This classification, alongside the language script, offers valuable insights for analyzing model behavior.

\begin{table}[htbp]
    \centering
    \resizebox{1\linewidth}{!}{
    \begin{tabular}{l}
        \toprule
        \makecell*[{{p{9.3cm}}}]{
        \makecell{\centering Simile CoT example of Swahili}\\
        \hline
        In this task, you are given a start phrase indicating a figurative expression in Swahili culture. Please select 0 if the start phrase conveys the meaning of ending 0, and 1 if it conveys the meaning of ending 1.\\
        \textbf{Start phrase:} He felt his chest was frozen like ice\\
        \textbf{Ending 0:} His heartbeat went too fast\\
        \textbf{Ending 1:} His heartbeat went slowly\\
        \textbf{Answer:} \blue{In Kiswahili culture}, \cyan{the expression "He felt his chest was frozen like ice" is used to show that a person felt fear, anxiety or uncertainty.} \orange{"The chest is frozen like ice" means that his emotions were stuck or frozen like ice. Therefore, the expression indicates a lack of enthusiasm. In this context, "His heartbeat was slow" indicates that the person's heartbeat was slow and steady. This indicates a sense of unhurriedness or excitement.} \green{Therefore, the answer is 1.}\\
        \textbf{<Test Simile Example>}\\
        \hline
        \makecell{\centering Idiom CoT example of German}\\
        \hline
        \textbf{Question:} How would one interpret this proverb in German culture, given the context? \purple{Please first think about the proverb’s meaning, then write an explanation of the proverb’s meaning}, and finally choose between A and B.\\
        \textbf{Proverb:} hope dies last\\
        \textbf{Context:} Person 1: Do you think we'll make it? Person 2: I don't know, but hope dies last.\\
        \textbf{Choices:} A: Person 2 hopes they'll make it. B: Person 2 has no hope they'll make it.\\
        \textbf{Explanation:} \purple{hope should be the last thing you give up}\\
        \textbf{Answer:} A}\\
        \textbf{<Test Idiom Example>}\\
        \bottomrule
    \end{tabular}
    }
    \caption{One-shot example of CoT prompts for simile and idiom. Thought pathways are specified with colors. \blue{blue}: specify culture, \cyan{cyan}: analyzing expression meaning, \orange{orange}: connect it to StartPhrase, \green{green}: specify the final answer. \purple{purple}: additional CoT trigger for idioms.}
    \label{tab:cot_one_shot_example_pathway}
\end{table}

\subsection{Models}
We evaluate the performance of several open-source and closed-source LLMs to understand their capabilities in processing figurative language across multiple languages. The models under consideration include GPT-3.5-Turbo-0125 \cite{openai2023gpt35}, GPT-4o mini \cite{openai2024gpt4technicalreport}, and Gemini 1.5 Flash \cite{geminiteam2024gemini15unlockingmultimodal} which are representative of closed-source commercial LLMs, as well as open-source models: Llama 3.1 \citep[8B, 70B]{dubey2024Llama3herdmodels} and Qwen2 \citep[7B, 72B]{yang2024qwen2technicalreport}.
These models are selected based on their widespread use and the contrast they offer in terms of accessibility, customization potential, and model size variations.
Note that although Llama 3.1 does not cover all of our languages (Just En, De, and Ru are covered), it performed quite well in our initial experiments.
The cost for running experiments is given in Appendix~\ref{appendix:experiment_cost}.

\subsection{Prompting}
We use several prompting techniques with examples in native or English translation or combined. We explain the techniques and then mention which ones we used in native, English, and native-English setups. The instruction is consistently given in English across all settings, as shown in Table~\ref{tab:cot_one_shot_example_pathway}. The full prompt templates used in the experiments are available in Appendix~\ref{appendix:prompt-templates}.
 
\subsubsection{Techniques}
We explore various prompt engineering techniques:
(i) \textbf{Zero-shot}: this setting assesses the model's basic understanding.
(ii) \textbf{One-shot}: here, we explore how providing a single example can enhance the model's knowledge of the task and its cultural context.
(iii) \textbf{Chain of Thought (CoT)}: this approach leverages the model's reasoning capabilities to break down and process figurative meanings step by step \citep{wei2023chainofthoughtpromptingelicitsreasoning},  which
is still effective under multilingual scenarios \cite{shi2022languagemodelsmultilingualchainofthought}. The instructions guide the model through a thought process to interpret idioms or similes. Additionally, a one-shot example is provided in the native language, accompanied by an explanation in the same language. 
For idioms, the one-shot example only explains the proverb's meaning. However, for similes, a reasoning pathway is provided that involves (1) mentioning the target culture, (2) interpreting the simile's meaning, (3) clarifying the reason for the similarity and its connection to the start phrase, and (4) generating the final answer. Examples of CoT are shown in Table~\ref{tab:cot_one_shot_example_pathway}.
(iv) \textbf{Dialogue simulation}: since figurative expressions often deviate from their literal meaning, understanding can be improved by placing them in context \citep{liu-etal-2024-multilingual}. We use the RiC (Reasoning in Conversation) method \citep{wang-etal-2024-reasoning} in zero-shot settings to prompt the model to generate dialogues or conversations between two individuals, embedding the figurative expression within these interactions.

\begin{table*}[!ht]
\resizebox{\textwidth}{!}{%
\begin{tabular}{@{}ccccccccccccccc@{}}
\cmidrule(l){2-15}
\multicolumn{1}{l}{}                           & \multicolumn{8}{c||}{Open Source}                                                                                                                                                    & \multicolumn{6}{c}{Closed Source}                                                                                                                                               \\ \midrule
\multicolumn{1}{c|}{\multirow{2}{*}{Language}} & \multicolumn{2}{c|}{Qwen2-7B}        & \multicolumn{2}{c|}{Qwen2-72B}                & \multicolumn{2}{c|}{Llama 3.1 8B}      & \multicolumn{2}{c||}{Llama 3.1 70B}                  & \multicolumn{2}{c|}{GPT-3.5}                                   & \multicolumn{2}{c|}{Gemini 1.5 Flash}                     & \multicolumn{2}{c}{GPT-4o mini}                     \\
\multicolumn{1}{c|}{}                          & Zero-Shot & \multicolumn{1}{c|}{CoT}  & Zero-Shot & \multicolumn{1}{c|}{CoT}           & Zero-Shot & \multicolumn{1}{c|}{CoT}  & Zero-Shot     & \multicolumn{1}{c||}{CoT}           & Zero-Shot                & \multicolumn{1}{c|}{CoT}           & Zero-Shot            & \multicolumn{1}{c|}{CoT}           & Zero-Shot                & CoT                      \\ \midrule \midrule
\multicolumn{15}{c}{\textbf{SIMILE}}                                                                                                                                                                                                                                                                                                                                                                                            \\ \midrule \midrule
\multicolumn{1}{c|}{En}                        & .651$_{.010}$      & \multicolumn{1}{c|}{.830$_{.010}$} & .936$_{.008}$      & \multicolumn{1}{c|}{\underline{\textbf{.943$_{.006}$}}} & .761$_{.030}$      & \multicolumn{1}{c|}{.770$_{.010}$} & .913$_{.008}$          & \multicolumn{1}{c||}{.883$_{.011}$}          & .786$_{.006}$                     & \multicolumn{1}{c|}{\underline{.916$_{.010}$}} & .878$_{.011}$                 & \multicolumn{1}{c|}{.896$_{.006}$}          & .740$_{.014}$                     & \underline{.916$_{.008}$}            \\
\multicolumn{1}{c|}{Id}                        & .590$_{.031}$      & \multicolumn{1}{c|}{.621$_{.035}$} & .893$_{.006}$      & \multicolumn{1}{c|}{\underline{.911$_{.009}$}} & .643$_{.027}$      & \multicolumn{1}{c|}{.728$_{.018}$} & .890$_{.010}$          & \multicolumn{1}{c||}{.895$_{.010}$}          & .683$_{.023}$                      & \multicolumn{1}{c|}{.805$_{.014}$}          & .910$_{.007}$                 & \multicolumn{1}{c|}{\underline{\textbf{.925$_{.004}$}}} & .711$_{.010}$                     & .911$_{.004}$                     \\
\multicolumn{1}{c|}{Hi}                        & .530$_{.018}$      & \multicolumn{1}{c|}{.583$_{.018}$} & .635$_{.010}$      & \multicolumn{1}{c|}{.638$_{.009}$}          & .533$_{.048}$      & \multicolumn{1}{c|}{.566$_{.004}$} & .655$_{.016}$          & \multicolumn{1}{c||}{\underline{.691$_{.010}$}} & .531$_{.013}$                     & \multicolumn{1}{c|}{.615$_{.010}$}          & .676$_{.004}$                 & \multicolumn{1}{c|}{\underline{\textbf{.740$_{.007}$}}} & .606$_{.022}$                     & .685$_{.008}$                     \\
\multicolumn{1}{c|}{Sw}                        & .508$_{.012}$      & \multicolumn{1}{c|}{.498$_{.013}$} & .618$_{.002}$      & \multicolumn{1}{c|}{.596$_{.040}$}          & .520$_{.017}$      & \multicolumn{1}{c|}{.523$_{.013}$} & \underline{.693$_{.026}$} & \multicolumn{1}{c||}{.681$_{.031}$}          & .533$_{.006}$                     & \multicolumn{1}{c|}{.714$_{.007}$}          & .745$_{.014}$                 & \multicolumn{1}{c|}{\underline{\textbf{.788$_{.012}$}}} & .556$_{.013}$                     & .768$_{.011}$                     \\
\multicolumn{1}{c|}{Jv}                        & .501$_{.008}$      & \multicolumn{1}{c|}{.536$_{.014}$} & .573$_{.015}$      & \multicolumn{1}{c|}{.656$_{.022}$}          & .501$_{.029}$      & \multicolumn{1}{c|}{.568$_{.019}$} & .669$_{.004}$ & \multicolumn{1}{c||}{.641$_{.020}$}          & .490$_{.008}$                     & \multicolumn{1}{c|}{.571$_{.020}$}          & \underline{.698$_{.010}$}        & \multicolumn{1}{c|}{.673$_{.009}$}          & .606$_{.009}$                     & \underline{\textbf{.783$_{.004}$}}                     \\
\multicolumn{1}{c|}{Kn}                        & .491$_{.023}$      & \multicolumn{1}{c|}{.493$_{.002}$} & .465$_{.014}$      & \multicolumn{1}{c|}{.514$_{.006}$}          & .403$_{.023}$      & \multicolumn{1}{c|}{.488$_{.009}$} & .493$_{.035}$          & \multicolumn{1}{c||}{\underline{.610$_{.008}$}} & .561$_{.002}$                     & \multicolumn{1}{c|}{.530$_{.003}$}          & .588$_{.004}$                 & \multicolumn{1}{c|}{\underline{\textbf{.619$_{.004}$}}} & .525$_{.020}$                     & .576$_{.006}$                     \\
\multicolumn{1}{c|}{Su}                        & .435$_{.022}$      & \multicolumn{1}{c|}{.505$_{.020}$} & .548$_{.025}$      & \multicolumn{1}{c|}{.568$_{.006}$}          & .456$_{.037}$      & \multicolumn{1}{c|}{.490$_{.008}$} & \underline{.593$_{.023}$} & \multicolumn{1}{c||}{.586$_{.048}$}          & .485$_{.004}$                     & \multicolumn{1}{c|}{.583$_{.004}$}          & .746$_{.010}$                 & \multicolumn{1}{c|}{.753$_{.009}$}          & .590$_{.014}$                     & \underline{\textbf{.768$_{.012}$}}            \\
\multicolumn{1}{c|}{Fa}                        & .576$_{.013}$         & \multicolumn{1}{c|}{.623$_{.009}$}    & .855$_{.010}$         & \multicolumn{1}{c|}{.864$_{.008}$}             & .658$_{.024}$         & \multicolumn{1}{c|}{.810$_{.020}$}    & .851$_{.014}$             & \multicolumn{1}{c||}{\underline{\textbf{.926$_{.015}$}}}             & .600$_{.017}$                    & \multicolumn{1}{c|}{.773$_{.006}$}             & \underline{{.915$_{.004}$}}                    & \multicolumn{1}{c|}{.830$_{.018}$}             & .680$_{.010}$                        & .898$_{.002}$                        \\ \midrule 
\multicolumn{1}{l|}{Average}                   & .535         & \multicolumn{1}{c|}{.586}    & .690         & \multicolumn{1}{c|}{.711}             & .559         & \multicolumn{1}{c|}{.618}    & .719             & \multicolumn{1}{c||}{\underline{{.739}}}            & .583                    & \multicolumn{1}{c|}{.688}             &  .769                    & \multicolumn{1}{c|}{.778}             & .627                        & \underline{\textbf{.788}}                        \\ \midrule \midrule
\multicolumn{15}{c}{\textbf{IDIOM}}                                                                                                                                                                                                                                                                                                                                                                                            \\ \midrule  \midrule
\multicolumn{1}{c|}{En}                        & .925$_{.003}$      & \multicolumn{1}{c|}{.915$_{.004}$} & .981$_{.002}$      & \multicolumn{1}{c|}{\underline{\textbf{.990$_{.004}$}}} & .878$_{.008}$      & \multicolumn{1}{c|}{.887$_{.004}$} & .975$_{.002}$          & \multicolumn{1}{c||}{.982$_{.004}$}          & .970$_{.004}$                     & \multicolumn{1}{c|}{.970$_{.002}$}          & .953$_{.000}$                 & \multicolumn{1}{c|}{.948$_{.006}$}          & .976$_{.005}$                     & \underline{.981$_{.004}$}            \\
\multicolumn{1}{c|}{Id}                        & .853$_{.008}$      & \multicolumn{1}{c|}{.831$_{.011}$} & .917$_{.003}$      & \multicolumn{1}{c|}{\underline{\textbf{.920$_{.004}$}}} & .711$_{.007}$      & \multicolumn{1}{c|}{.801$_{.002}$} & .895$_{.015}$          & \multicolumn{1}{c||}{.914$_{.008}$}          & .852$_{.007}$                     & \multicolumn{1}{c|}{.789$_{.009}$}          & .900$_{.004}$                 & \multicolumn{1}{c|}{.912$_{.009}$}          & \underline{.918$_{.008}$}            & .894$_{.007}$                     \\
\multicolumn{1}{c|}{Zh}                        & .878$_{.009}$      & \multicolumn{1}{c|}{.871$_{.006}$} & .979$_{.003}$      & \multicolumn{1}{c|}{\underline{\textbf{.986$_{.003}$}}} & .815$_{.010}$      & \multicolumn{1}{c|}{.767$_{.008}$} & .942$_{.003}$          & \multicolumn{1}{c||}{.953$_{.009}$}          & .874$_{.000}$                     & \multicolumn{1}{c|}{.920$_{.014}$}          & \underline{.979$_{.000}$}        & \multicolumn{1}{c|}{.951$_{.009}$}          & .921$_{.006}$                     & .944$_{.011}$                     \\
\multicolumn{1}{c|}{Bn}                        & .659$_{.006}$      & \multicolumn{1}{c|}{.643$_{.004}$} & .896$_{.002}$      & \multicolumn{1}{c|}{\underline{\textbf{.913$_{.004}$}}} & .663$_{.010}$      & \multicolumn{1}{c|}{.649$_{.008}$} & .838$_{.016}$          & \multicolumn{1}{c||}{.855$_{.014}$}          & .510$_{.008}$                     & \multicolumn{1}{c|}{.611$_{.009}$}          & .861$_{.001}$                 & \multicolumn{1}{c|}{.849$_{.009}$}          & .845$_{.010}$                     & \underline{.872$_{.007}$}            \\
\multicolumn{1}{c|}{Ru}                        & .845$_{.003}$      & \multicolumn{1}{c|}{.823$_{.004}$} & .933$_{.002}$      & \multicolumn{1}{c|}{\underline{\textbf{.938$_{.006}$}}} & .778$_{.007}$      & \multicolumn{1}{c|}{.805$_{.012}$} & .914$_{.014}$          & \multicolumn{1}{c||}{.917$_{.004}$}          & .838$_{.005}$                     & \multicolumn{1}{c|}{.849$_{.009}$}          & \underline{.914$_{.007}$}        & \multicolumn{1}{c|}{.910$_{.007}$}          & .877$_{.004}$                     & .874$_{.004}$                     \\
\multicolumn{1}{c|}{De}                        & .862$_{.002}$      & \multicolumn{1}{c|}{.830$_{.006}$} & .939$_{.003}$      & \multicolumn{1}{c|}{.950$_{.004}$}          & .759$_{.009}$      & \multicolumn{1}{c|}{.803$_{.006}$} & .943$_{.003}$          & \multicolumn{1}{c||}{\underline{\textbf{.957$_{.004}$}}} & .877$_{.002}$                     & \multicolumn{1}{c|}{.901$_{.004}$}          & .894$_{.009}$                 & \multicolumn{1}{c|}{.878$_{.005}$}          & .901$_{.002}$                     & \underline{.911$_{.006}$}            \\
\multicolumn{1}{c|}{Fa}                        & .841$_{.012}$      & \multicolumn{1}{c|}{.788$_{.010}$} & .943$_{.010}$      & \multicolumn{1}{c|}{\underline{.955$_{.007}$}} & .756$_{.005}$      & \multicolumn{1}{c|}{.813$_{.009}$} & .924$_{.018}$          & \multicolumn{1}{c||}{.941$_{.012}$}          & .828$_{.004}$                     & \multicolumn{1}{c|}{.853$_{.010}$}          & \underline{\textbf{.964$_{.001}$}}                    & \multicolumn{1}{c|}{.943$_{.002}$}             & .940$_{.009}$                     & .936$_{.009}$                     \\ \midrule
\multicolumn{1}{l|}{Average}                   & .837      & \multicolumn{1}{c|}{.814} & .941      & \multicolumn{1}{c|}{\underline{\textbf{.950}}} & .765      & \multicolumn{1}{c|}{.789} & .918          & \multicolumn{1}{c||}{.931}          & \multicolumn{1}{c}{.821} & \multicolumn{1}{l|}{.841}          & \multicolumn{1}{c}{\underline{.923}} & \multicolumn{1}{c|}{.913}              & \multicolumn{1}{c}{.911} & \multicolumn{1}{l}{.916} \\ \bottomrule
\end{tabular}%
}
\caption{Results of open-source and closed-source LLMs with native input languages, averaged over three runs (std is also shown). The best accuracy for each category of models is \underline{underlined}, and the best overall accuracy is \textbf{bold}.}
\label{tab:open-close-models-native-comparison}
\end{table*}

\subsubsection{Input language}
\label{sec:input_lang}
\paragraph{Native} 
In this setting, the input examples are presented in their original language (e.g., Chinese, Indonesian, etc.) along with instructions in English. We evaluate our experiments across three different configurations: zero-shot, one-shot, and CoT.

\paragraph{Translated-English} 
We follow the trend of translating either the training or test data into English \citep{shi2023language, conneau-etal-2018-xnli, qin-etal-2023-cross}, utilizing three translation systems that cover all of our languages: the Google Translate API, Meta’s No Language Left Behind (NLLB-200-3.3B) model \citep{nllbteam2022languageleftbehindscaling}, and GPT-3.5 \cite{openai2023gpt35}. We investigate zero-shot and one-shot settings to explore the few-shot effect in this context. For CoT, we ask the model to explain the meanings of idioms or similes alongside a one-shot example in English to activate CoT reasoning abilities. For simile, we also experiment with dialogue simulation prompting techniques. For idiom tasks, as they are already presented in a dialogue format, dialogue simulation is not applied.

\paragraph{Native and Translated-English} In this approach, we directly prompt the model to convert the input from the native language to English and then perform the task \cite{etxaniz-etal-2024-multilingual, huang-etal-2023-languages}. The input is provided in both native and Translated-English simultaneously, leveraging the model’s intrinsic translation capabilities. This technique is applied within a one-shot setting, as detailed in Tables~\ref{tab:prompt_temp_self_translate_mabl} and ~\ref{tab:prompt_temp_self_translate_maps}.

\subsection{Evaluation Process} We utilize LLMs to evaluate simile and idiom tasks. For each task, we select an LLM, apply the appropriate prompt, and retrieve the model’s answer. The answers are parsed using regular expressions (regex) to extract the final binary result. In cases where regex fails to correctly extract the answer due to irregular formatting or model output variations, manual verification is employed to ensure accuracy in the evaluation process.

\begin{figure*}[!ht]
    \centering
    \begin{subfigure}[b]{1\linewidth}
        \centering
        \includegraphics[width=1\linewidth]{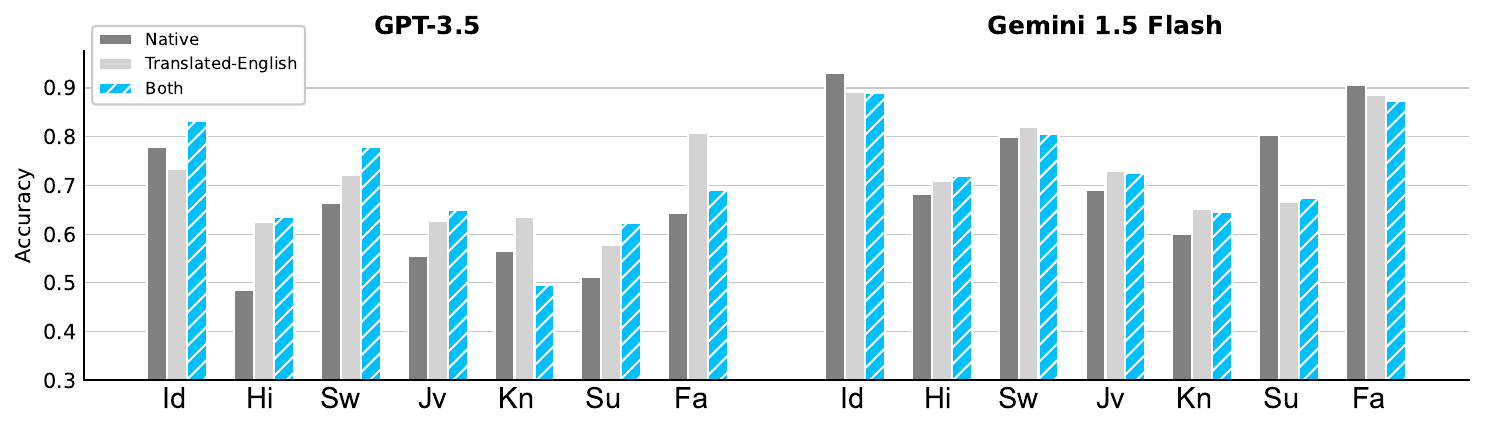}
        \caption{Simile}
        \label{fig:native_translate_compare_models_simile}
    \end{subfigure}
    
    \begin{subfigure}[b]{1\linewidth}
        \centering
        \includegraphics[width=1\linewidth]{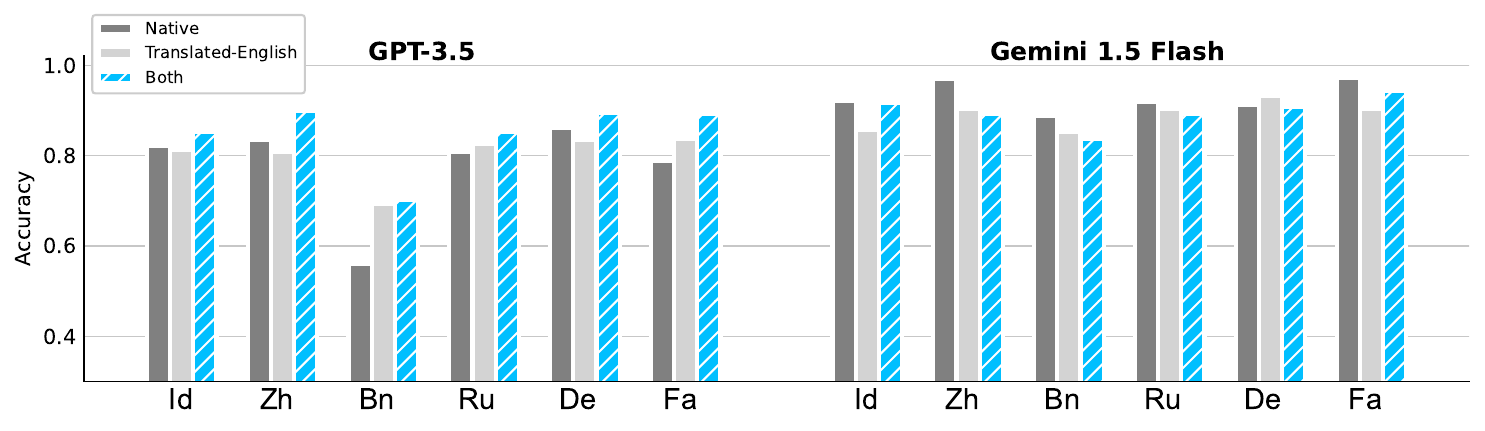}
        \caption{Idiom}
        \label{fig:native_translate_compare_models_idiom}
    \end{subfigure}
    
    \caption{Comparing results of inputs being in native, Translated-English, and both languages in one-shot setting.}
    \label{fig:native_translate_compare_models}
\end{figure*}

\section{Results} In this section, we present the performance of various large language models (LLMs) across two datasets and multiple languages.

\subsection{Native} We evaluate the performance of LLMs on input provided in native languages. Results for closed- and open-source models are shown in Table~\ref{tab:open-close-models-native-comparison}.
The results show that both types of models can comprehend similes and idioms to varying extents. 
Generally, performance on idioms is higher than similes. 

For similes, open-source models with fewer parameters show lower accuracies, with Llama 3.1-8B outperforming Qwen2-7B. Large open-source models, i.e., Llama 3.1-70B and Qwen2-72B, outperform GPT-3.5. Among all models, however, Gemini 1.5 demonstrates the best simile interpretation performance on average across our two prompting strategies.
Notably, Gemini 1.5 significantly boosts zero-shot performance in Sundanese (a low-resource language), achieving up to 73\% accuracy, while the second-best model remains below 60\%.

For idioms, the open-source model Qwen2-72B outperforms other models, even the closed-source ones. Interestingly, Qwen2, with only 7B parameters, outperforms GPT-3.5 in zero-shot on average, especially in Bengali and Persian. Among closed-source models, Gemini 1.5 remains the best. Notably, Gemini 1.5 demonstrates significant performance in Chinese and Persian, achieving accuracy levels even higher than in English, which highlights its strong capability in handling these specific languages compared with other models. 

Overall, the results reveal that model performance varies significantly depending on the type of figurative language being evaluated. Moreover, they highlight that in understanding figurative language, an open-source model can outperform closed-source ones in specific languages. This suggests that success in comprehending figurative expressions may depend more on specific language or cultural knowledge encoded in the models, which might need further exploration in future work. 

\paragraph{Impact of model size and CoT}
Since we do not have access to the size of closed-source models,
we consider only open-source models when analyzing the effect of 
model size here.
We observe that in nearly all cases, increasing the model size improves accuracy for both simile and idiom tasks. The effect is particularly pronounced in simile interpretation, where performance improves by about 16\% absolute point on average. For specific languages such as English and Indonesian, the improvement reaches around 30\%. In contrast, the improvement for idiom tasks is less significant on average.

Our analysis reveals that the effectiveness of \textbf{CoT} prompting varies by model size. For similes, smaller models, such as Qwen2-7B and Llama 3.1 8B, exhibit greater error reduction with CoT prompting, with reductions of approximately 11\% and 13\%, respectively. In comparison, larger models like Qwen2-72B and Llama 3.1 70B show smaller but still notable reductions, each around 7\%. This suggests that CoT is particularly beneficial for smaller models in simile interpretation, likely by enhancing their reasoning capabilities.

For idioms, however, the pattern is different. While Llama 3.1 8B achieves a modest error reduction of 10\%, Qwen2-7B experiences a performance drop of 14\%, indicating that smaller models may struggle with the complexity of idiom interpretation when CoT is applied. Conversely, larger models like Qwen2-72B and Llama 3.1 70B demonstrate strong error reductions of over 15\% each. These findings suggest that while larger models benefit from CoT across both tasks, their already strong reasoning capabilities make CoT less impactful for simile interpretation but more crucial for idiomatic expressions. Similar trends have been observed in other studies, such as \citet{sprague2024cotcotchainofthoughthelps}, where the impact of CoT was found to be more pronounced in smaller open-source models than in larger ones.

On the other hand for closed-source models, CoT compared to zero-shot has improved efficiency for GPT-4o mini and GPT-3.5, though this trend is not consistently observed for Gemini 1.5. Additionally, CoT tends to be more effective for similes.

\paragraph{Cross-lingual interference in smaller models} 
During our experiments, we observe that smaller models, particularly Llama 3.1 8B, occasionally exhibit cross-lingual interference. For instance, while generating responses in Persian, the model sometimes inserts Chinese characters or words in the middle of the response, only to revert back to Persian in the continuation of the text. This inconsistency suggests that smaller models may struggle to maintain coherence in the intended language, potentially due to confusion in multilingual settings.

\subsection{English translation}
\label{sec:translated-english}
Figure~\ref{fig:native_translate_compare_models} represents our evaluations on GPT-3.5 as our base model and Gemini 1.5 Flash as the strongest closed-source model. We experiment with native and Translated-English inputs in the one-shot setting. Google Translate is used for our translation. Other translation methods are examined in Section \ref{sec:translation_methods}

\paragraph{Comparing Native with Translated-English}
In both figurative tasks, GPT-3.5 exhibits lower accuracy across all languages when compared to Gemini 1.5, a trend also evident in native prompting.
The use of translation has made a significant improvement in the performance of GPT-3.5 for similes. We can also see this trend with Gemini 1.5, but it is not consistent for all languages. With Gemini 1.5, native prompting surpasses translation for Indonesian, Persian, and Sundanese, which may show the superiority of the Gemini model in understanding these languages in their native format.
When it comes to idioms, Gemini 1.5 performs slightly better in native prompts, though the improvement is minor. GPT-3.5, however, shows varying results depending on the language. For lower resource and non-Latin languages like Bengali, Persian, and Russian, GPT-3.5 tends to achieve better results when using Translated-English rather than native prompts, likely due to its weaknesses in handling these languages. Generally, it can be concluded that translation efficiency depends on the language, LLM, and task. This observation aligns with the conclusions drawn in the study by \citet{zhang2023m3exammultilingualmultimodalmultilevel}, suggesting that translation's effectiveness depends on whether potential comprehension gains outweigh translation errors, and it may not always enhance performance.

\paragraph{Input in Native and Translated-English}
The results of using both native and English inputs (as explained in Section \ref{sec:input_lang}) are shown in Figure~\ref{fig:native_translate_compare_models} in the third column named ``Both''. 
When the GPT-3.5 model is used, using both native and Translated-English is a more effective method in most languages in both figurative types. There are two exceptions in similes for Kannada and Persian, where the Translated-English approach performs better. This seems to be specific to certain languages and requires further investigation. 
When using Gemini 1.5, a more capable model, the results are similar to those obtained using the Translated-English method for both figurative types. We also observe that Gemini 1.5 generally does not outperform native prompting in idioms. This indicates the model's strong understanding of idioms in native, and translating them may lead to errors in interpretation.

\subsubsection{Comparing translation methods}
\label{sec:translation_methods}
\begin{figure}[!ht]
    \centering
    \begin{subfigure}[b]{1\linewidth}
        \centering
        \includegraphics[width=1\linewidth]{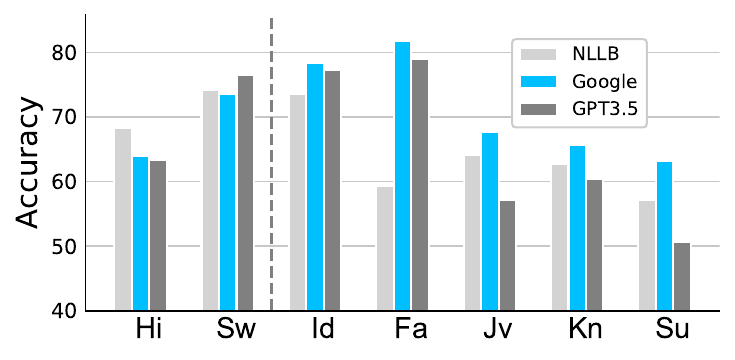}
        \caption{Simile}
        \label{fig:TT_Compare_simile}
    \end{subfigure}

    \begin{subfigure}[b]{1\linewidth}
        \centering
        \includegraphics[width=1\linewidth]{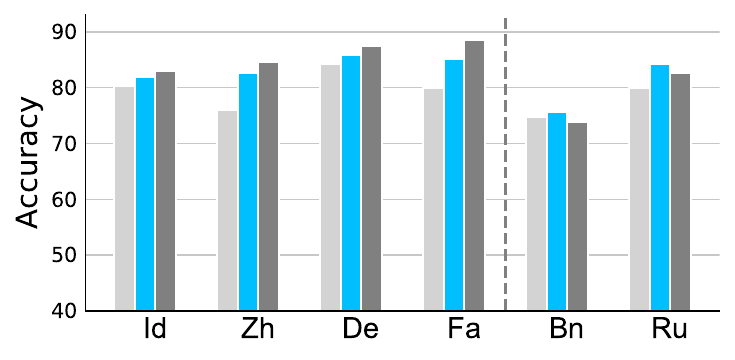}
        \caption{Idiom}
        \label{fig:TT_Compare_idiom}
    \end{subfigure}
    
    \caption{Comparing the translation methods using GPT-3.5. The average accuracy of zero-shot, one-shot, and CoT methods is reported for each translation method.}
    \label{fig:TT_Compare_Both}
\end{figure}

So far in the reported results, the translation is done using only Google Translate. Here, we investigate two additional methods (NLLB and GPT-3.5) to translate our datasets, followed by evaluations in zero-shot, one-shot, and CoT settings for GPT-3.5. To streamline our conclusions, we report the average performance of these prompting techniques across languages and figurative types in Figure~\ref{fig:TT_Compare_Both}.

For idioms, GPT-3.5 outperforms both Google Translate and NLLB for most high-resource or Latin languages (e.g., In, Zh, De, and Fa).
However, for similes, Google and NLLB perform better than GPT-3.5 in lowest-resource languages (Jv, Kn, Su). This result aligns with our expectations, as these models were trained to provide better translations for a wider range of languages, including low-resource ones. This finding also mirrors the results of \citet{liu2024translationneedstudysolving}, which highlight that while NLLB shows strong performance, Google Translate tends to outperform it in most scenarios when handling multilingual tasks.
Overall, translation with Google is the better choice for similes, while GPT-3.5 excels with idioms. This distinction may arise from the fact that literal translations by Google are more likely to alter the meaning of idioms compared to similes.

\section{Analysis}
In this section, we present further analyses of our observations to deepen our comparative study of simile and idiom interpretation. 

\subsection{The thought pathway in CoT}
We investigate the thought process and pathway of the LLMs doing CoT in our experiments. 
We analyze instances where models initially generate incorrect answers in a zero-shot setting but provide accurate responses when using CoT. We focus on closed-source models here.

For idiom interpretation, we observe that models predominantly generate responses and explanations in English, even when the one-shot example is presented in a different native language. In contrast, for simile interpretation, the behavior varies across models: GPT-4o mini consistently responds in the native language, while GPT-3.5 produces a mix of English and native language responses, with the ratio of English outputs varying depending on the language. Gemini 1.5, however, tends to generate explanations primarily in English, which can lead to misinterpretations of culturally specific concepts. Detailed results of language detection in CoT responses are provided in Table~\ref{tab:lang_detection_results}.

To further investigate the impact of language in CoT responses on simile interpretation, we conducted additional experiments. We explicitly prompted Gemini 1.5 to generate responses in the native language, and instructed GPT-4o mini to respond in English (Appendix~\ref{sec:influence_of_cot_lang}). While GPT-4o mini showed no significant performance differences when responding in English (except Sw), Gemini 1.5 displayed improved performance in Fa, Su, and Jv when responding in their respective native languages. Results from these new experiments are presented in Table~\ref{tab:new_CoT_prompts} in the Appendix.

\subsection{Dialogue simulation}
We evaluate the dialogue simulation technique for simile interpretation, expecting context to improve understanding \cite{liu-etal-2024-multilingual}. However, results show that CoT prompting outperforms dialogue simulation across almost all languages (Table~\ref{tab:ric}).
This shows that reasoning and then explicitly deriving the meaning of a simile (as in the one-shot CoT setting) is a better approach than using it in a context like a conversation.

\begin{table}[ht]
    \centering
    \resizebox{7.5cm}{!}{
        \begin{tabular}{c|ccccccc}
        lang. & Id & Hi & Sw & Jv & Kn & Su & Fa \\
        \hline
        CoT & \textbf{.864} & \textbf{.685} & \textbf{.818} & \textbf{.735} & \textbf{.705} & .678 & \textbf{.855} \\
        Dial. & .803 & .661 & .751 & .698 & .538 & \textbf{.696} & .806 \\
        \end{tabular}
    }
    \caption{Comparison of GPT-3.5 performance using CoT and dialogue simulation techniques in translated-English prompts. *Dial. refers to dialogue simulation.}
    \label{tab:ric}
\end{table}

\subsection{Consistency of results}
We examine the reliability of some prompting methods in answering questions with \textit{consistency} metric, i.e. the proportion of samples where the model gives the same answer across all three runs, regardless of whether the answer is correct or not. 

\begin{figure}
    \centering
    \includegraphics[width=1\linewidth]{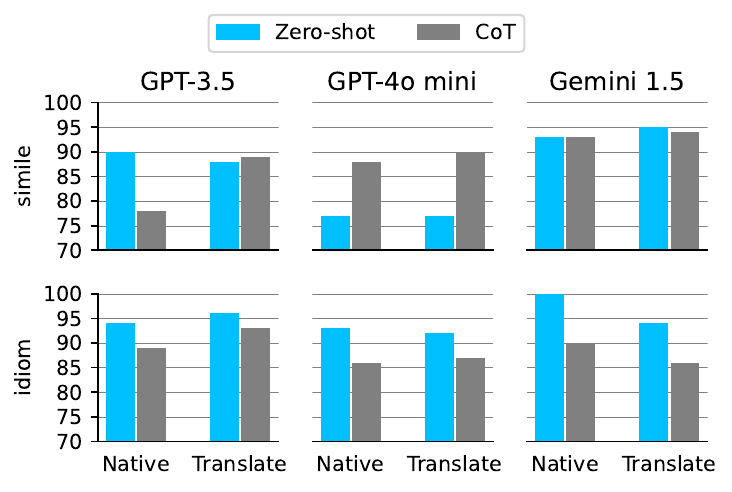}
    \caption{Comparing consistency (\%) of closed-source models in zero-shot and CoT settings when input is in native or Translated-English. Each number represents the average on languages.}
    \label{fig:closed_source_consistencies}
\end{figure}

We employ the consistency metric to show the uncertainty and reliability of the models. By examining how consistency shifts from zero-shot to CoT in idiom, we observe that models generally exhibit increased uncertainty when they generate an explanation before providing an answer. However, the effect of CoT varies across different models for simile. Specifically, we find that GPT-4o mini demonstrates greater reliability after using CoT, while GPT-3.5 in native prompting shows a decrease. In contrast, Gemini 1.5 maintains a steady level of consistency throughout.

\subsection{Making idiom examples more challenging}
To better understand the limitations of the datasets better, we conduct an analysis on the Persian idiom dataset. We modify 60 samples by replacing one of the answer choices with the literal meaning of the idiom while retaining the correct non-literal meaning. This approach tests the model’s ability to distinguish between figurative and literal interpretations.
For instance, as shown in Table~\ref{tab:analysis-4.4}, the idiom ``\small \FR{آب از دست کسی نچکیدن}'' \normalsize which literally means ``water does not drip from someones hand'', is a figurative expression meaning ``someone is very stingy''. 
The results indicate an 8\% absolute decrease in accuracy for GPT-4o mini (i.e., a drop in 4 out of 60 examples), underscoring the model’s challenges in accurately interpreting idiomatic expressions when presented with plausible literal alternatives. This finding emphasizes the inherent complexity of figurative language understanding for LLMs.

\section{Conclusion}
This study examined how multilingual LLMs interpret similes and idioms across languages, comparing open-source and closed-source models using MABL, MAPS, and newly developed Persian datasets. We tested prompt engineering strategies like one-shot, CoT, and dialogue simulation, finding their effectiveness varied by figurative language type, language, and model. While open-source models like Llama 3.1 and Qwen2 performed well overall, they struggled with similes in low-resource languages. Idiom interpretation, however, showed near-saturation, highlighting the need for more challenging datasets.

Our two new Persian datasets contribute valuable resources for evaluating LLMs in this language. Expanding the scope of figurative language types to include metaphors, sarcasm, and irony could provide a more comprehensive evaluation of LLMs' capabilities. Also, developing datasets that challenge LLMs with more context-dependent or ambiguous figurative expressions will be crucial for driving progress in this area.

\section{Limitation}
While our study offers valuable insights into multilingual figurative language understanding, several limitations remain. First, we primarily focus on similes and idioms, excluding other important figurative types like metaphors, sarcasm, and irony due to the scarcity of relevant datasets. Additionally, the datasets used in this research cover different languages, with only English and Indonesian being common across both, complicating cross-language comparisons. The dataset quality, especially for low-resource languages, also could not be verified by native speakers, potentially introducing inaccuracies in culturally specific expressions. Moreover, many open-source models, including Llama 3.1 and Qwen2, lack official support for low-resource languages like Sundanese and Javanese, making their performance in these contexts less reliable. Lastly, the datasets may not be challenging enough for advanced models, as GPT-4o mini and Gemini 1.5 Flash nearly achieve perfect accuracy in high-resource languages, pointing to the need for more complex and context-dependent figurative tasks to fully test model capabilities.

\section*{Acknowledgements}
We extend our gratitude to Yong Zheng-Xin and Emmy Liu for their valuable feedback and insightful comments on an earlier version of this paper. We also sincerely thank the anonymous reviewers for their valuable feedback and thoughtful suggestions.

\bibliography{custom}

\appendix
\section{Simile and Idiom Datasets}
\label{appendix:datasets_intro}
Here is a detailed explanation of the MABL and MAPS datasets, designed to evaluate figurative language understanding for similes and idioms, respectively.

\subsection{MABL for Simile}
The MABL dataset is designed to assess figurative language understanding, consists of similes. Similes are rhetorical devices that compare two different concepts, often using words like "as" or "like", to highlight similarities. Understanding similes requires a nuanced grasp of both literal and figurative meanings within a sentence, making this a challenging task for language models.

The primary task in the MABL dataset is a binary classification problem. Each instance, as shown in Table~\ref{tab:mabl_dataset}, consists of:
\textbf{start phrase}, a sentence containing a simile,
Two possible continuations, \textbf{ending 0} and \textbf{ending 1}, where one reflects the correct figurative meaning and the other is often opposing or incorrect, and
A binary \textbf{label} indicating the correct ending.
This task evaluates whether language models can correctly interpret the intended figurative meaning of similes.

\subsection{MAPS for Idiom}
The MAPS dataset is designed to evaluate the understanding and interpretation of proverbs and idioms within conversational contexts. Proverbs and idioms are commonly used expressions that convey figurative meanings, which often differ significantly from their literal interpretations. Accurate comprehension requires models to infer the intended figurative meaning of these expressions based on context.

The MAPS dataset also presents a binary classification problem. Each instance, as illustrated in Table~\ref{tab:maps_dataset}, consists of:
\textbf{proverb}, a commonly used saying or idiomatic expression, \textbf{conversation}, a short dialogue where the proverb or idiom is used, two possible interpretations of the given conversation, one aligned with the figurative meaning of the proverb, and the other aligned with a literal or incorrect interpretation, and binary \textbf{label} (A or B) indicating the correct interpretation.
This task assesses whether language models can accurately interpret the figurative meaning of idioms and proverbs when presented within a conversational context.

\section{Dataset Construction}
\label{appendix:dataset-construction}
In this section, we provide additional details on the construction of the Persian simile and idiom datasets.

\subsection{Persian Simile Dataset} To create the Persian Simile dataset, we task GPT-4o with generating over 800 simile pairs, following the structure used in the MABL dataset. Each sample consists of a start phrase with two possible endings: one correctly reflecting the intended meaning and the other presenting an incorrect or opposite interpretation. Table~\ref{tab:simile_data_construction_prompt} provides one example of used prompts.

\begin{table}[!htb]
    \centering
    \resizebox{0.88\linewidth}{!}{
    \begin{tabular}{l}
        \toprule
        \makecell*[{{p{7.8cm}}}]{
        \textcolor{black}{Your task is to generate pairs of sentences with opposite or very different meanings, both of which contain Persian similes. You can feel free to incorporate creativity into the similes, but also make sure that they’re something that could be understood by the speakers of the language that you are generating similes for, e.g., “this is as classic as pancakes for breakfast” to mean “this is classic” wouldn’t make sense for a culture in which pancakes aren’t traditionally eaten for breakfast. You can do this by thinking of a simile that conveys a certain meaning, and replacing the similative phrase with another similative phrase of the same type that conveys the opposite meaning. Here are some examples of similes to give you an idea of what we’re looking for: Please write the simile and two correct and incorrect meanings.}\\
        \small \FR{1. دست‌هایش مثل تنه‌ی درخت بلوط، قوی و محکم هستند.}\\
        \textcolor{gray}{(Her hands are as strong and sturdy as the trunk of an oak tree.)}\\
        \small \FR{دست‌هایش ضعیف و بی‌رمق هستند / دست‌هایش قوی و پرتوان هستند}\\
        \textcolor{gray}{(Her hands are weak and frail / Her hands are strong and powerful)}\\
        \small \FR{۲. رفتارش مثل نسیمی است که در میان گل‌ها می‌وزد}\\
        \textcolor{gray}{(Her behavior is like a breeze blowing among the flowers)}\\
        \small \FR{رفتارش پرخاشگرانه و تند است./رفتارش آرام و متین است.}\\
        \textcolor{gray}{(Her behavior is aggressive and abrupt / His behavior is calm and composed)}
        }\\
        \bottomrule
    \end{tabular}
    }
    \caption{An example of the prompt used in the construction of the Persian simile dataset.}
    \label{tab:simile_data_construction_prompt}
\end{table}

\paragraph{Sample Refinement.} After generation, two native Persian speakers review all the samples, focusing on grammatical accuracy, cultural relevance, and the quality of the figurative language. From the 800+ generated samples, the reviewers select approximately 200 of the most relevant and high-quality examples. They further analyze the selected samples, refining those that require adjustments. The refinement process includes correcting figurative meanings and incorporating cultural references and specific characters to enhance the dataset's complexity and relevance for Persian speakers. The reviewers validate both the correctness of the simile usage and the plausibility of the incorrect endings, ensuring that they are clearly distinguishable yet realistic. 

The final Persian Simile dataset contains 200 samples, each validated for correctness and cultural appropriateness, making it a robust resource for evaluating simile comprehension in Persian.

\subsection{Persian Idiom Dataset} To create the Persian Idiom dataset, we first collect over 400 idioms along with their meanings from two online resources, Daneshchi and Abadis. From these, we remove any idioms deemed inappropriate or unsuitable for our purposes. We then use GPT-4o to generate conversational contexts for the remaining idioms, ensuring that the idioms are correctly integrated within natural dialogues.

\paragraph{Sample Refinement.} Two native Persian speakers review the generated conversational contexts, refining them to ensure grammatical accuracy, cultural relevance, and correct idiomatic usage. During this process, we incorporate cultural references, including historical characters, cultural concepts, and significant events—such as the great wars in the history of Iran—to increase the dataset's complexity and make it more representative of Persian culture, similar to the approach used in the MABL dataset.

\paragraph{Option Generation.} In the second round, GPT-4o generates two response options for each conversational context: one correct and one incorrect, based on the meaning of the idiom within the dialogue. Once again, two native Persian speakers review the generated responses, ensuring that the grammar, idiom usage, and cultural aspects are accurate and appropriate.

The final Persian Idiom dataset contains 316 samples, each with a conversational context and two response options. It serves as a valuable resource for evaluating idiomatic comprehension in Persian, emphasizing cultural and linguistic accuracy.

\section{Cost For Running Experiments}
\label{appendix:experiment_cost}
We access these models through the APIs provided by OpenAI \footnote{\href{https://platform.openai.com/}{https://platform.openai.com}}, Gemini \footnote{\href{https://ai.google.dev/}{https://ai.google.dev}}, and OpenRouter \footnote{\href{https://openrouter.ai/}{https://openrouter.ai}} for accessing open-source models. The total cost for running the experiments is estimated to be under \$40, with approximately \$30 spent on OpenAI and \$10 on OpenRouter.

\section{Prompt Templates}
\label{appendix:prompt-templates}
In this section, we provide the prompting techniques templates used for different tasks, such as zero-shot, one-shot, and dialogue simulation (RiC).

\subsection{Zero-shot}
This template is designed to assess the model’s basic understanding without providing any examples. The instructions are in English, while the input may be in the native language or English. The prompt templates for MABL and MAPS samples are shown in Tables~\ref{tab:prompt_temp_zero_mabl} and~\ref{tab:prompt_temp_zero_maps}.

\begin{table}[htbp]
    \centering
    \resizebox{0.88\linewidth}{!}{
    \begin{tabular}{l}
        \toprule
        \makecell*[{{p{8.1cm}}}]{
        \textcolor{black}In this task, you are given a start phrase indicating a figurative expression in <language> culture. Please select 0 if the start phrase conveys the meaning of ending 0, and 1 if it conveys the meaning of ending 1.\\ \\
        \textcolor{black}{\textbf{Start Phrase:}} <start phrase> \\
        \textcolor{black}{\textbf{Ending 0:}} <ending 0>\\
        \textcolor{black}{\textbf{Ending 1:}} <ending 1>\\
        \textcolor{black}{\textbf{Answer:}}\\
        }\\
        \bottomrule
    \end{tabular}
    }
    \caption{Zero-shot prompt template for MABL samples.}
    \label{tab:prompt_temp_zero_mabl}
\end{table}

\begin{table}[htbp]
    \centering
    \resizebox{0.88\linewidth}{!}{
    \begin{tabular}{l}
        \toprule
        \makecell*[{{p{7.8cm}}}]{
        \textcolor{black}{\textbf{Question:} How would one interpret this proverb in <language> culture, given the context? Please choose between A and B.}\\
        \textcolor{black}{\textbf{Proverb:}} <proverb> \\
        \textcolor{black}{\textbf{Context:}} <context>\\
        \textcolor{black}{\textbf{Choices:}} A: <answer A> B: <answer B> \\
        \textcolor{black}{\textbf{Answer:}}
        }\\
        \bottomrule
    \end{tabular}
    }
    \caption{Zero-shot prompt template for MAPS samples.}
    \label{tab:prompt_temp_zero_maps}
\end{table}

\subsection{One-shot} 
In the one-shot template, a single example is provided to guide the model in understanding the task. The instruction is given in English, while the input language can vary between native and translated-English (as shown in Tables~\ref{tab:prompt_temp_one-mabl} and~\ref{tab:prompt_temp_one-maps}), or a combination of both native and translated-English (Tables~\ref{tab:prompt_temp_self_translate_mabl} and~\ref{tab:prompt_temp_self_translate_maps}).

\begin{table}[htbp]
    \centering
    \resizebox{0.88\linewidth}{!}{
    \begin{tabular}{l}
        \toprule
        \makecell*[{{p{8.1cm}}}]{
        \textcolor{black}In this task, you are given a start phrase indicating a figurative expression in <language> culture. Please select 0 if the start phrase conveys the meaning of ending 0, and 1 if it conveys the meaning of ending 1.\\ \\
        Below is an example showing you how to do the task: \\
        \textcolor{black}{\textbf{Start Phrase:}} <sample start phrase> \\
        \textcolor{black}{\textbf{Ending 0:}} <sample ending 0>\\
        \textcolor{black}{\textbf{Ending 1:}} <sample ending 1>\\ 
        \textcolor{black}{\textbf{Answer:}} <0/1>\\ \\
        Now answer the following question:\\ 
        \textcolor{black}{\textbf{Start Phrase:}} <start phrase> \\
        \textcolor{black}{\textbf{Ending 0:}} <ending 0>\\
        \textcolor{black}{\textbf{Ending 1:}} <ending 1>\\ 
        \textcolor{black}{\textbf{Answer:}}
        }\\
        \bottomrule
    \end{tabular}
    }
    \caption{One-shot prompt template for MABL samples (Native or Translated-English).}
    \label{tab:prompt_temp_one-mabl}
\end{table}

\begin{table}[htbp]
    \centering
    \resizebox{0.88\linewidth}{!}{
    \begin{tabular}{l}
        \toprule
        \makecell*[{{p{8.5cm}}}]{
        \textcolor{black}{\textbf{Question:} How would one interpret this proverb in <language> culture, given the context? Please choose between A and B.}\\
        \textcolor{black}{\textbf{Proverb:}} <sample proverb> \\
        \textcolor{black}{\textbf{Context:}} <sample context> \\
        \textcolor{black}{\textbf{Choices:}} A: <sample answer A> B: <sample answer B> \\
        \textcolor{black}{\textbf{Answer:}} <A/B> \\ \\
        \textcolor{black}{\textbf{Question:} How would one interpret this proverb in <language> culture, given the context? Please choose between A and B.}\\
        \textcolor{black}{\textbf{Proverb:}} <proverb> \\
        \textcolor{black}{\textbf{Context:}} <context>\\
        \textcolor{black}{\textbf{Choices:}} A: <answer A> B: <answer B> \\
        \textcolor{black}{\textbf{Answer:}}
        }\\
        \bottomrule
    \end{tabular}
    }
    \caption{One-shot prompt template for MAPS samples (Native or Translated-English).}
    \label{tab:prompt_temp_one-maps}
\end{table}

\begin{table}[htbp]
    \centering
    \resizebox{0.88\linewidth}{!}{
    \begin{tabular}{l}
        \toprule
        \makecell*[{{p{8.4cm}}}]{
        \textcolor{black}In this task, you are given a start phrase indicating a figurative expression in <language> culture. Please first translate the start phrase, ending 0, and ending 1 into English. Then, select 0 if the translated start phrase conveys the meaning of the translated Ending 0, and 1 if it conveys the meaning of the translated Ending 1.\\ \\
        Below is an example showing you how to do the task: \\
        \textcolor{black}{\textbf{Start Phrase:}} <sample start phrase> \\
        \textcolor{black}{\textbf{Ending 0:}} <sample ending 0>\\
        \textcolor{black}{\textbf{Ending 1:}} <sample ending 1>\\ \\ 
        \textcolor{black}{\textbf{Translated into English:}} \\
        \textcolor{black}{\textbf{Start Phrase:}} <sample start phrase English translation> \\
        \textcolor{black}{\textbf{Ending 0:}} <sample ending 0 English translation>\\
        \textcolor{black}{\textbf{Ending 1:}} <sample ending 1 English translation>\\
        \textcolor{black}{\textbf{Answer:}} <0/1>\\ \\
        Now answer the following question:\\ 
        \textcolor{black}{\textbf{Start Phrase:}} <start phrase> \\
        \textcolor{black}{\textbf{Ending 0:}} <ending 0>\\
        \textcolor{black}{\textbf{Ending 1:}} <ending 1>\\ 
        \textcolor{black}{\textbf{Translated into English:}} \\
        \textcolor{black}{\textbf{Answer:}}
        }\\
        \bottomrule
    \end{tabular}
    }
    \caption{One-shot prompt template for MABL samples (Native and Translated-English).}
    \label{tab:prompt_temp_self_translate_mabl}
\end{table}

\begin{table}[htbp]
    \centering
    \resizebox{0.88\linewidth}{!}{
    \begin{tabular}{l}
        \toprule
        \makecell*[{{p{8.5cm}}}]{
        \textcolor{black}{\textbf{Question:} How would one interpret this proverb in \textless language\textgreater\ culture, given the context? Please first translate the Proverb, Context, and Choices into English. Then, choose between A and B.}\\
        \textcolor{black}{\textbf{Proverb:}} <sample proverb> \\
        \textcolor{black}{\textbf{Context:}} <sample context>\\
        \textcolor{black}{\textbf{Choices:}} A: <sample answer A> B: <sample answer B> \\ \\
        \textcolor{black}{\textbf{The English translation is:}}\\
        \textcolor{black}{\textbf{Proverb:}} <sample proverb English translation> \\
        \textcolor{black}{\textbf{Context:}} <sample context English translation>\\
        \textcolor{black}{\textbf{Choices:}} A: <sample answer A English translation> B: <sample answer B English translation> \\
        \textcolor{black}{\textbf{Final Answer:}} <A/B>\\ \\
        \textcolor{black}{\textbf{Question:} How would one interpret this proverb in \textless language\textgreater\ culture, given the context? Please first translate the Proverb, Context, and Choices into English. Then, choose between A and B.}\\
        \textcolor{black}{\textbf{Proverb:}} \textless proverb\textgreater \\
        \textcolor{black}{\textbf{Context:}} \textless context\textgreater\\
        \textcolor{black}{\textbf{Choices:}} A: <answer A> B: <answer B> \\ \\
        \textcolor{black}{\textbf{The English translation is:}}\\
        \textcolor{black}{\textbf{Final Answer:}}
        }\\
        \bottomrule
    \end{tabular}
    }
    \caption{{One-shot prompt template for MAPS samples (Native and Translated-English).}}
    \label{tab:prompt_temp_self_translate_maps}
\end{table}

\subsection{Chain of Thought (CoT)}
This template is designed to test the model’s ability to reason through the task step-by-step before arriving at a final decision. The prompt includes an example that demonstrates how the model should generate both an answer and the reasoning behind it. The CoT prompt structure for similes and idioms are detailed in Table~\ref{tab:prompt_temp_cot_mabl} and Table~\ref{tab:prompt_temp_cot_maps} respectively.

\begin{table}[htb]
    \centering
    \resizebox{0.88\linewidth}{!}{
    \begin{tabular}{l}
        \toprule
        \makecell*[{{p{8.1cm}}}]{
        \textcolor{black} In this task, you are given a start phrase indicating a figurative expression in <language> culture. Please select 0 if the start phrase conveys the meaning of ending 0, and 1 if it conveys the meaning of ending 1.\\ \\
        Below is an example showing you how to do the task: \\
        \textcolor{black}{\textbf{Start Phrase:}} <sample start phrase> \\
        \textcolor{black}{\textbf{Ending 0:}} <sample ending 0>\\
        \textcolor{black}{\textbf{Ending 1:}} <sample ending 1>\\
        \textcolor{black}{\textbf{Answer:}} <sample answer with reasoning>\\ \\    
        \textcolor{black} Now answer the following question:\\ 
        \textcolor{black}{\textbf{Start Phrase:}} <start phrase> \\
        \textcolor{black}{\textbf{Ending 0:}} <ending 0>\\
        \textcolor{black}{\textbf{Ending 1:}} <ending 1>\\
        \textcolor{black}{\textbf{Answer:}}
        }\\
        \bottomrule
    \end{tabular}
    }
    \caption{CoT prompt template for MABL samples.}
    \label{tab:prompt_temp_cot_mabl}
\end{table}

\begin{table}[htb]
    \centering
    \resizebox{0.88\linewidth}{!}{
    \begin{tabular}{l}
        \toprule
        \makecell*[{{p{8.7cm}}}]{
        \textcolor{black}{\textbf{Question:} How would one interpret this proverb in <language> culture, given the context? Please first think about the proverb's meaning, then write an explanation of the proverb's meaning, and finally choose between A and B.}\\
        \textcolor{black}{\textbf{Proverb:}} <sample proverb> \\
        \textcolor{black}{\textbf{Context:}} <sample context> \\
        \textcolor{black}{\textbf{Choices:}} A: <sample answer A> B: <sample answer B> \\
        \textcolor{black}{\textbf{Explanation:}} <sample explanation> \\
        \textcolor{black}{\textbf{Answer:}} <A/B>\\ \\    
        \textcolor{black}{\textbf{Question:} How would one interpret this proverb in <language> culture, given the context? Please first think about the proverb's meaning, then write an explanation of the proverb's meaning, and finally choose between A and B.}\\
        \textcolor{black}{\textbf{Proverb:}} <proverb> \\
        \textcolor{black}{\textbf{Context:}} <context>\\
        \textcolor{black}{\textbf{Choices:}} A: <answer A> B: <answer B> \\
        \textcolor{black}{\textbf{Explanation:}} \\ 
        \textcolor{black}{\textbf{Answer:}}
        }\\
        \bottomrule
    \end{tabular}
    }
    \caption{CoT prompt template for MAPS samples.}
    \label{tab:prompt_temp_cot_maps}
\end{table}

\subsection{Dialogue Simulation (RiC)}
This template is used in the RiC method, where the model generates a conversation that includes the figurative expression, helping it understand the phrase in context. The prompt template for MABL samples is shown in Table~\ref{tab:prompt_temp_ric}.

\begin{table}[htbp]
    \centering
    \resizebox{0.88\linewidth}{!}{
    \begin{tabular}{l}
        \toprule
        \makecell*[{{p{7.8cm}}}]{
        {\textbf{Figurative Language Interpretation:}} In this task, you are given a start phrase indicating a figurative expression in <language> culture. Please select 0 if the start phrase conveys the meaning of ending 0, and 1 if it conveys the meaning of ending 1.\\ \\
        {\textbf{Start Phrase:}} <start phrase> \\
        {\textbf{Ending 0:}} <ending 0>\\
        {\textbf{Ending 1:}} <ending 1>\\ \\
        First, extract keywords from the question. \\
        Then, according to the keywords, construct a scenario for the question in the form of dialogue. \\
        Finally, according to the question and conversation, reason and give the final answer. Select from 0 or 1.\\
        }\\
        \bottomrule
    \end{tabular}
    }
    \caption{Dialogue simulation (RiC) prompt template for MABL samples.}
    \label{tab:prompt_temp_ric}
\end{table}

\section{Language detection of CoT responses of closed-source models}
We use the Google Translate Python library for language detection of CoT responses in the native prompting setting for closed-source models in both simile and idiom. Results are in Table~\ref{tab:lang_detection_results}.

\begin{table}[htb]
    \centering
    \begin{tabular}{|l|c|ccc|}
        \hline
        & S & GPT-4o mini & GPT3.5 & Gemini 1.5 \\
        \hline \hline
        &&\multicolumn{3}{|c|}{Simile} \\
        \hline \hline
        En & 200 & 100 & 100 & 100 \\
        Hi & 200 & 0 & 3.5 & 91.4 \\
        Id & 200 & 0 & 10 & 45.5 \\
        Sw & 200 & 0 & 4.5 & 98.9 \\
        Jv & 200 & 5.5 & 46.5 & 93.2 \\
        Kn & 200 & 2 & 35.7 & 92.1 \\
        Su & 200 & 0 & 0.5 & 88.9 \\
        Fa & 200 & 0 & 13 & 51.4 \\
        \hline \hline
        &&\multicolumn{3}{|c|}{Idiom} \\
        \hline \hline
        En & 206 & 100 & 100 & 100 \\
        Zh & 139 & 96.5 & 95.1 & 100 \\
        Bn & 262 & 93.7 & 24.7 & 95 \\
        Ru & 213 & 99.6 & 65.5 & 98.6 \\
        De & 172 & 95.1 & 99.5 & 100 \\
        Id & 248 & 93.6 & 45.3 & 99.6 \\
        Fa & 299 & 74.1 & 13 & 88.3 \\
        \hline
    \end{tabular}
    \caption{Ratio (\%) of CoT examples in which their languages are detected as English. Examples of native prompting in the CoT setting are considered. S refers to the number of samples of each language.}
    \label{tab:lang_detection_results}
\end{table}

\section{Examine the influence of CoT responses language}
\label{sec:influence_of_cot_lang}
As the Gemini 1.5 and GPT-4o mini exhibit different behaviors in the language used for generating responses for CoT, two additional experiments were conducted to examine the influence of language used in the responses. In the new experiments, the phrase "\textit{Give the Answer in <language> language}" was appended to the end of the prompts. For Gemini, the model was asked to respond in its native language of example, while GPT-4o was instructed to respond in English. The accuracy results for the new prompts are presented in Table~\ref{tab:new_CoT_prompts}.

\begin{table}[!ht]
    \centering
    \resizebox{1\linewidth}{!}{
    \begin{tabular}{c|cc|cc}
        \multirow{2}{*}{Lang.} & \multicolumn{2}{c|}{Gemini 1.5} & \multicolumn{2}{c}{GPT-4o mini}\\
         & \makecell{Default\\(English)} & \makecell{New\\(Native)} & \makecell{Default\\(Native)} & \makecell{New\\(English)} \\
        \hline
        En & .887 & .900 & .916 & .925 \\
        Id & .924 & .910 & .911 & .913 \\
        Hi & .744 & .695 & .685 & .696 \\
        Sw & .804 & .775 & .768 & .803 \\
        Jv & .670 & .760 & .783 & .794 \\
        Kn & .622 & .585 & .576 & .580 \\
        Su & .745 & .790 & .768 & .766 \\
        Fa & .825 & .925 & .898 & .896 \\
        \hline
        Avg & .777 & .792 & .788 & .797 \\
    \end{tabular}
    }
    \caption{Performance (\%) of Gemini 1.5 when prompted to respond in native, and GPT-4o minie when prompted to respond in English. Experiments are on the simile dataset for native prompting and CoT setting.}
    \label{tab:new_CoT_prompts}
\end{table}

\section{Distinguish Between Figurative and Literal Meaning of Idioms}
Here is an example of the prompt structure in Table~\ref{tab:analysis-4.4}.
\begin{table}[!htb]
    \centering
    \resizebox{0.88\linewidth}{!}{
    \begin{tabular}{l}
        \toprule
        \makecell*[{{p{7.8cm}}}]{
        \textcolor{black}{\textbf{Question:} What is the meaning of the following phrase in Persian, given the context? Please choose between A and B.}\\
        {\textbf{Phrase:}} \small \FR{آب از دست کسی نچکیدن} \\ 
        \textcolor{gray}{(Water does not drip from someone's hand)} \\
        \textcolor{black}{\textbf{Context:}} \small \FR{شخص ۱: از اون خواستم مقداری پول به من قرض دهد. شخص ۲: اون را می‌شناسم. آب از دستش نمی‌چکد!} \\ 
        \textcolor{gray}{(Person 1: I asked them to lend me some money. Person 2: I know them. Water does not drip from their hand!)} \\
        \textcolor{black}{\textbf{Choices:}}\\
        \textcolor{black}{\textbf{A:} \small \FR{فردی خسیس و تنگ نظر بودن}} \\
        \textcolor{gray}{(Being stingy and narrow-minded)} \\
        \textcolor{black}{\textbf{B:} \small \FR{محافظ خوبی برای آب بودن}} \\ 
        \textcolor{gray}{(Being a good guardian of water)} \\
        \textcolor{black}{\textbf{Answer:}}\\
        }\\
        \bottomrule
    \end{tabular}
    }
    \caption{Prompt structure used for analyzing the model's ability to distinguish between figurative and literal meanings. Choice A indicates figurative meaning, while Choice B provides a plausible literal interpretation. Note: English translations, provided in parentheses below the original Persian phrases, are not part of the prompt presented to the model.}
    \label{tab:analysis-4.4}
\end{table}

\end{document}